\def\year{2018}\relax
\documentclass[letterpaper]{article} %DO NOT CHANGE THIS
\usepackage{aaai18}  %Required
\usepackage{times}  %Required
\usepackage{helvet}  %Required
\usepackage{courier}  %Required
\usepackage{url}  %Required
\usepackage{graphicx}  %Required

\nocopyright

\begin{document}
	\title{PixelLink: Detecting Scene Text via Instance Segmentation}
	\author{Dan Deng$^{1,3}$\thanks{Part of this work was done when Dan Deng was an intern at Visual Computing Group, CVTE Research.}, \quad Haifeng Liu$^{1}$, \quad Xuelong Li$^{4}$, \quad Deng Cai$^{1,2}$ \\
		$^{1}$State Key Lab of CAD\&CG, College of Computer Science, Zhejiang University\\
		$^{2}$Alibaba-Zhejiang University Joint Institute of Frontier Technologies\\
		$^{3}$CVTE Research\\
		$^{4}$Xi'an Institute of Optics and Precision Mechanics, Chinese Academy of Sciences \\
		dengdan.zju@gmail.com \quad  \{haifengliu,dcai\}@zju.edu.cn \quad  xuelong\_li@opt.ac.cn\\
	}
	\maketitle
	\begin{abstract}
		Most state-of-the-art scene text detection algorithms are deep learning based methods that depend on bounding box regression and perform at least two kinds of predictions: text/non-text classification and location regression. Regression plays a key role in the acquisition of bounding boxes in these methods, but it is not indispensable because text/non-text prediction can also be considered as a kind of semantic segmentation that contains full location information in itself. However, text instances in scene images often lie very close to each other, making them very difficult to separate via semantic segmentation. Therefore, instance segmentation is needed to address this problem. In this paper, PixelLink, aƒ novel scene text detection algorithm based on instance segmentation, is proposed. Text instances are first segmented out by linking pixels within the same instance together. Text bounding boxes are then extracted directly from the segmentation result without location regression. Experiments show that, compared with regression-based methods, PixelLink can achieve better or comparable performance on several benchmarks, while requiring many fewer training iterations and less training data.
	\end{abstract}
	
	\section{Introduction}
	% text detection task
	Reading text in the wild, or robust reading has drawn great interest for a long time~\cite{Ye2015Survey}. It is usually divided into two steps or sub-tasks: text detection and text recognition.
	
	The detection task, also called localization, takes an image as input and outputs the locations of text within it. Along with the advances in deep learning and general object detection, more and more accurate as well as efficient scene text detection algorithms have been proposed, \emph{e.g.}, CTPN~\cite{Tian2016CTPN}, TextBoxes~\cite{Liao2016TextBoxes}, SegLink~\cite{Shi2017SegLink} and EAST~\cite{Zhou2017EAST}.
	Most of these state-of-the-art methods are built on Fully Convolutional Networks~\cite{Long2015FCN}, and perform at least two kinds of predictions:
	\begin{enumerate}
		\item Text/non-text classification. Such predictions can be taken as probabilities of pixels being within text bounding boxes \cite{zhang2016TextBlock}. But they are more frequently used as confidences on regression results (\emph{e.g.}, TextBoxes, SegLink, EAST).
		
		\item Location regression. Locations of text instances, or their segments/slices, are predicted as offsets from reference boxes (\emph{e.g.}, TextBoxes, SegLink, CTPN), or absolute locations of bounding boxes (\emph{e.g.}, EAST).
	\end{enumerate}
	In methods like SegLink, linkages between segments are also predicted.  After these predictions, post-processing that mainly includes joining segments together (\emph{e.g.}, SegLink, CTPN) or Non-Maximum Suppression (\emph{e.g.}, TextBoxes, EAST), is applied to obtain bounding boxes as the final output.
	
	Location regression has long been used in object detection, as well as in text detection, and has proven to be effective. It plays a key role in the formulation of text bounding boxes in state-of-the-art methods. However, as mentioned above, text/non-text predictions can not only be used as the confidences on regression results, but also as a segmentation score map, which contains location information in itself and can be used to obtain bounding boxes directly. Therefore, regression is not indispensable.
	
\begin{figure}
		\centering
		\includegraphics[width=0.95\linewidth]{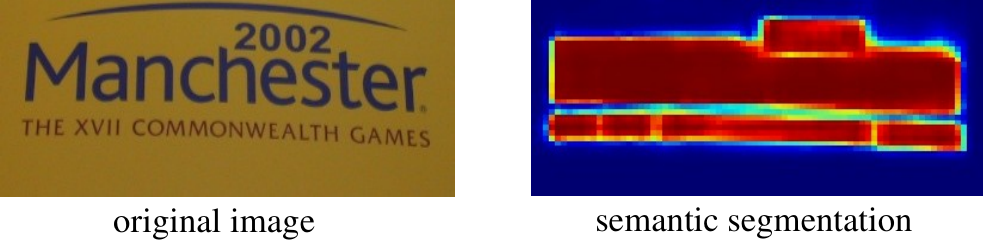}
		\label{figure:problem1}
		\caption{Text instances often lie close to each other, making them hard to separate via semantic segmentation.}
	\end{figure}

	However, as shown in Fig.~\ref{figure:problem1}, text instances in scene images usually lie very close to each other. In such cases, they are very difficult, and are sometimes even impossible to separate via semantic segmentation (\emph{i.e.}, text/non-text prediction) only; therefore, segmentation at the instance level is further required.
	
	% overall introduction to pixel link
	To solve this problem, a novel scene text detection algorithm, PixelLink, is proposed in this paper. It extracts text locations directly from an instance segmentation result, instead of from bounding box regression. In PixelLink, a Deep Neural Network (DNN) is trained to do two kinds of pixel-wise predictions, text/non-text prediction, and link prediction. Pixels within text instances are labeled as positive (\emph{i.e.}, text pixels), and otherwise are labeled as negative (\emph{i.e.}, non-text pixels). The concept of link here is inspired by the link design in SegLink, but with significant difference. Every pixel has 8 neighbors. For a given pixel and one of its neighbors, if they lie within the same instance, the link between them is labeled as positive, and otherwise negative. Predicted positive pixels are joined together into Connected Components (CC) by predicted positive links. Instance segmentation is achieved in this way, with each CC representing a detected text. Methods like \emph{minAreaRect} in OpenCV~\cite{opencv} can be applied to obtain the bounding boxes of CCs as the final detection result.
	
	Our experiments demonstrate the advantages of PixelLink over state-of-the-art methods based on regression. Specifically, trained from scratch, PixelLink models can achieve comparable or better performance on several benchmarks while requiring fewer training iterations and less training data.
	
	\section{Related Work}
	\subsection{Semantic\&Instance Segmentation}
	The segmentation task is to assigning pixel-wise labels to an image. When only object category is considered, it is called semantic segmentation. Dominating methods for this task usually adopts the approach of Fully Convolution Networks (FCN)~\cite{Long2015FCN}. Instance segmentation is more challenging than semantic segmentation because it requires not only object category for each pixel but also a differentiation of instances. It's more relevant to general object detection than semantic segmentation, for being aware of object instances. Recent methods in this field make heavy use of object detection systems. FCIS~\cite{Li2016FCIS} extends the idea of position-sensitive prediction in R-FCN~\cite{dai2016rfcn}. Mask R-CNN~\cite{He2017MaskRCNN} changes the RoIPooling in Faster R-CNN~\cite{Ren2017Faster} to RoIAlign. They both do detection and segmentation in a same deep model, and highly depend their segmentation results on detection performance.
	
	\subsection{Segmentation-based Text Detection}
	Segmentation has been adopted in text detection for a long time. \cite{yao2016scene} cast the detection task as a semantic segmentation problem, by predicting three kinds of score maps: text/non-text, character classes, and character linking orientations. They are then grouped into words or lines. In ~\cite{zhang2016TextBlock}, TextBlocks are found from a saliency map predicted by FCN, and character candidates are extracted using MSER~\cite{Donoser2006MSER}. Lines or words are formed using hand-crafted rules at last. In CCTN~\cite{He2016CCTN}, a coarse network is used to detect text regions roughly by generating a text region heat-map, and then the detected regions are refined into text lines by a fine text network, which outputs a central line area heat-map and a text line area heat-map. These methods often suffer from time-consuming post-processing steps and unsatisfying performances.
	
	\subsection{Regression-based Text Detection}
	Most methods in this category take advantage of the development in general object detection. CTPN~\cite{Tian2016CTPN} extends the anchor idea in object detection to predict text slices, which are then connected through heuristic rules. TextBoxes~\cite{Liao2016TextBoxes}, a text-specific SSD~\cite{Liu2016SSD}, adopts anchors of large aspect ratio and kernels of irregular shape, to fit for the large-aspect-ratio feature of scene text.  RRPN~\cite{Ma2017Arbitrary} adds rotation to both anchors and RoIPooling in Faster R-CNN, to deal with the orientation of scene text. SegLink~\cite{Shi2017SegLink} adopts SSD to predict text segments, which are linked into complete instances using the linkage prediction.  EAST~\cite{Zhou2017EAST} performs very dense predictions that are processed using locality-aware NMS. All these regression-based text detection algorithms have predictions for both confidences and locations at the same time.
	
	In this paper, state-of-the-art mainly refers to published methods that perform best on IC13~\cite{Karatzas2013ICDAR} or IC15~\cite{Karatzas2015ICDAR}, including TextBoxes, CTPN, SegLink, and EAST.
	
	\section{Detecting Text via Instance Segmentation}
\begin{figure*}[!h]
		\begin{center}
			\includegraphics[width=0.9\linewidth]{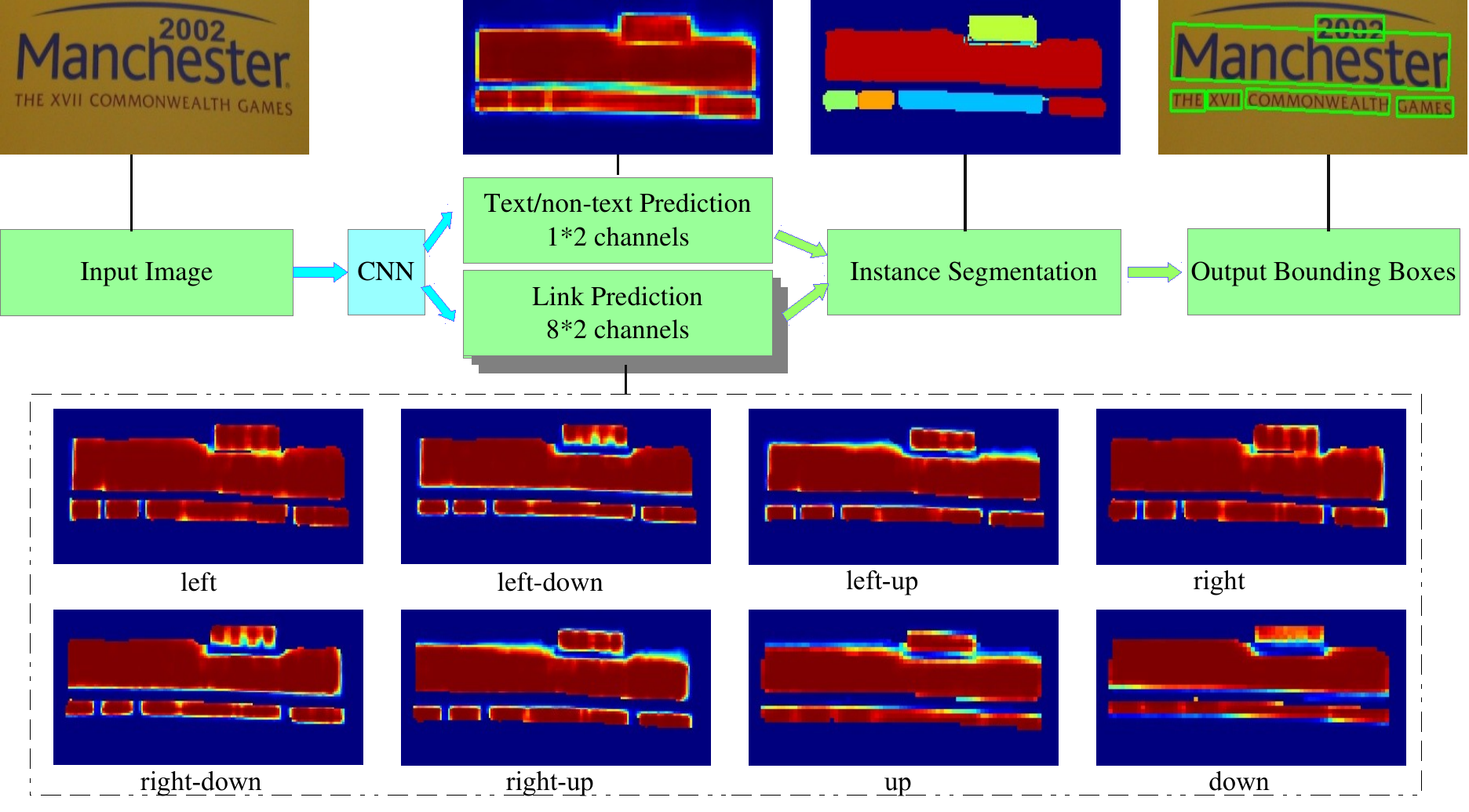}
		\end{center}
		\caption{Architecture of PixelLink. A CNN model is trained to perform two kinds of pixel-wise predictions: text/non-text prediction and link prediction. After being thresholded, positive pixels are joined together by positive links, achieving instance segmentation. \emph{minAreaRect} is then applied to extract bounding boxes directly from the segmentation result. Noise predictions can be efficiently removed using post-filtering. An input sample is shown for better illustration. The eight heat-maps in the dashed box stand for the link predictions in eight directions. Although some words are difficult to separate in text/non-text prediction, they are separable through link predictions.}
		\label{fig:alg-demo}
	\end{figure*}

	As shown in Fig.~\ref{fig:alg-demo}, PixelLink detects text via instance segmentation, where predicted positive pixels are joined together into text instances by predicted positive links. Bounding boxes are then directly extracted from this segmentation result.
	\subsection{Network Architecture}
\begin{figure}[t!]
		\centering
		\includegraphics[width=0.9\linewidth]{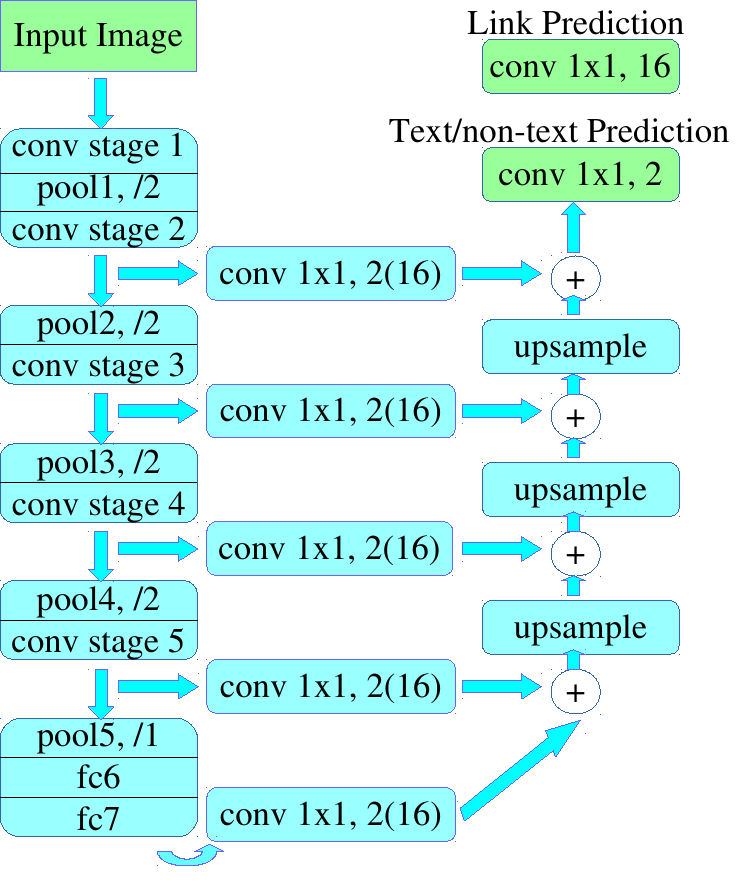}
		\caption{Structure of PixelLink+VGG16 2s.
			\emph{fc6 }and \emph{fc7} are converted into convolutional layers. The upsampling operation is done through bilinear interpolation directly. Feature maps from different stages are fused through a cascade of upsampling and add operations. All pooling layers except \emph{pool5} take a stride of 2, and \emph{pool5} takes 1. Therefore, the size of \emph{fc7} is the same as \emph{conv5\_3}, and no upsampling is needed when adding scores from these two layers. \mbox{`\emph{conv $1\times1$,2(16)}'} stands for a $1\times 1$ convolutional layer with 2 or 16 kernels, for text/non-text prediction or link prediction individually.}
		\label{figure:vgg16}
	\end{figure}

	Following SSD and SegLink, VGG16~\cite{simonyan2014VGG} is used as the feature extractor, with fully connected layers, \emph{i.e.}, \emph{fc6} and \emph{fc7}, being converted into convolutional layers. The fashion of feature fusion and pixel-wise prediction inherits from ~\cite{Long2015FCN}. As shown in Fig.~\ref{figure:vgg16}, the whole model has two separate headers, one for text/non-text prediction, and the other for link prediction. Softmax is used in both, so their outputs have 1*2=2 and 8*2=16 channels, respectively.
	
	Two settings of feature fusion layers are implemented: \emph{\{conv2\_2, conv3\_3, conv4\_3, conv5\_3, fc\_7\}}, and \emph{\{conv3\_3, conv4\_3, conv5\_3, fc\_7\}}, denoted as \textbf{PixelLink+VGG16 2s}, and \textbf{PixelLink+VGG16 4s}, respectively. The resolution of 2s predictions is a half of the original image, and 4s is a quarter.
	
	\subsection{Linking Pixels Together}
	\label{sec:linking-pixels-together}
	Given predictions on pixels and links, two different thresholds can be applied  on them separately. Positive pixels are then grouped together using positive links, resulting in a collection of CCs, each representing a detected text instance. Thus instance segmentation is achieved.  It is worth noting that, given two neighboring positive pixels, their link are predicted by both of them, and they should be connected when one or both of the two link predictions are positive. This linking process can be implemented using disjoint-set data structure.

	\subsection{Extraction of Bounding Boxes}
	Actually, the detection task is completed after instance segmentation. However, bounding boxes are required as detection results in challenges like IC13~\cite{Karatzas2013ICDAR}, IC15~\cite{Karatzas2015ICDAR}, and COCO-Text~\cite{Veit2016COCO}. Therefore, bounding boxes of CCs are then extracted through methods like \emph{minAreaRect} in OpenCV~\cite{opencv}. The output of \emph{minAreaRect} is an oriented rectangle, which can be easily converted into quadrangles for IC15, or rectangles for IC13. It is worth mentioning that in PixelLink, there is no restriction on the orientation of scene text.
	
	This step leads to the key difference between PixelLink and regression-based methods, \emph{i.e.}, bounding boxes are obtained directly from instance segmentation other than location regression.
	
	\subsection{Post Filtering after Segmentation}
	Since PixelLink attempts to group pixels together via links, it is inevitable to have some noise predictions, so a post-filtering step is necessary. A straightforward yet efficient solution is to filter via simple geometry features of detected boxes, \emph{e.g.}, width, height, area and aspect ratio, \emph{etc.}  For example, in the IC15 experiments in Sec.~\ref{sec:results-on-ic15}, a detected box is abandoned if its shorter side is less than 10 pixels or if its area is smaller than 300. The 10 and 300 are statistical results on the training data of IC15. Specifically, for a chosen filtering criteria, the corresponding 99-th percentile calculated on \textbf{TRAINING} set is chosen as the threshold value. For example, again, 10 is chosen as the threshold on shorter side length because about 99\% text instances in IC15-train have a shorter side $\ge 10$ pixels.
	
	\section{Optimization}
	\subsection{Ground Truth Calculation}
	Following the formulation in TextBlocks~\cite{zhang2016TextBlock}, pixels inside text bounding boxes are labeled as positive. If overlapping exists, only un-overlapped pixels are positive. Otherwise negative.
	
	For a given pixel and one of its eight neighbors, if they belong to the same instance, the link between them is positive. Otherwise negative.
	
	Note that ground truth calculation is carried out on input images resized to the shape of prediction layer, \emph{i.e.}, \emph{conv3\_3} for 4s and \emph{conv2\_2} for 2s.
	
	\subsection{Loss Function}
	The training loss is a weighted sum of loss on pixels and loss on links:
	\begin{equation}
	L = \lambda L_{pixel} + L_{link}.
	\end{equation}
	Since $L_{link}$ is calculated on positive pixels only, the classification task of pixel is more important than that of link,  and $\lambda$ is set to 2.0 in all experiments.
	\subsubsection{Loss on Pixels}
	Sizes of text instances might vary a lot. For example, in Fig.~\ref{figure:problem1}, the area of `Manchester' is greater than the sum of all the other words. When calculating loss, if we put the same weight on all positive pixels, it's unfair to instances with small areas, and may hurt the performance. To deal with this problem, a novel weighted loss for segmentation, \emph{Instance-Balanced Cross-Entropy Loss}, is proposed.
	In detail, for a given image with $N$ text instances, all instances are treated equally by giving a same weight to everyone of them, denoted as $B_i$ in Equ.~\ref{equ:loss-per-instance}. For the $i$-th instance with $area=S_i$, every positive pixels within it have a weight of $  w_i = \frac{B_i}{S_i}$.
	\begin{equation}
	B_i = \frac {S}{N}, S = \sum_i^N S_i, \forall i \in \{1, \dots, N\} \label{equ:loss-per-instance}
	\end{equation}

	Online Hard Example Mining (OHEM)~\cite{Shrivastava2016OHEM} is applied to select negative pixels. More specifically, $r*S$ negative pixels with the highest losses are selected, by setting their weights to ones. ~$r$ is the negative-positive ratio and is set to 3 as a common practice.
	
	The above two mechanisms result in a weight matrix, denoted by $W$, for all positive pixels and selected negative ones. The loss on pixel classification task is:
	\begin{equation}
	L_{pixel} = \frac{1}{(1 + r)S} W L_{pixel\_CE}
	\label{equ:loss-on-pixel-classification},
	\end{equation}
	where $L_{pixel\_CE}$ is the matrix of Cross-Entropy loss on text/non-text prediction.
	
	As a result, pixels in small instances have a higher weight, and pixels in large instances have a smaller weight. However, every instance contributes equally to the loss.
	
	\subsubsection{Loss on Links}
	Losses for positive and negative links are calculated separately and on positive pixels only:
	$$
	L_{link\_pos} = W_{pos\_link} L_{link\_CE},
	$$
	$$
	L_{link\_neg}= W_{neg\_link} L_{link\_CE},
	$$
	where $L_{link\_CE}$ is the Cross-Entropy loss matrix on link prediction.  $W_{pos\_link}$ and $W_{neg\_link}$ are the weights of positive and negative links respectively. They are calculated from the $W$ in Equ.~\ref{equ:loss-on-pixel-classification}.
	In detail, for the $k$-th neighbor of pixel $(i, j)$:
	$$
	W_{pos\_link}(i,j,k) = W(i, j) * (Y_{link}(i,j,k) == 1), 
	$$
	$$
	W_{neg\_link}(i,j,k) = W(i, j) * (Y_{link}(i,j,k) == 0), 
	$$
	where $Y_{link}$ is the label matrix of links.\\
	
	The loss on link prediction is a kind of class-balanced cross-entropy loss:
	\begin{equation}
	L_{link} = \frac{L_{link\_pos}}{rsum(W_{pos\_link})} + \frac{L_{link\_neg}}{rsum(W_{neg\_link})}, 
	\end{equation}
	where $rsum$ denotes \emph{reduce sum}, which sums all elements of a tensor into scalar.
	
	\subsection{Data Augmentation}
	Data augmentation is done in a similar way to SSD with an additional random rotation step. Input images are firstly rotated at a probability of 0.2, by a random angle of $0, \pi / 2, \pi$, or $3\pi/2$, the same with~\cite{He2017DDR}. Then randomly crop them with areas ranging from 0.1 to 1, and aspect ratios ranging from 0.5 to 2. At last, resize them uniformly to $512 \times 512$.
	After augmentation, text instances with a shorter side less than 10 pixels are ignored. Text instances remaining less than $20\%$ are also ignored. Weights for ignored instances are set to zero during loss calculation.
	
	\section{Experiments}
	\label{sec:benchmark-results}
\begin{figure*}
		\begin{center}
			\includegraphics[width=0.9\linewidth]{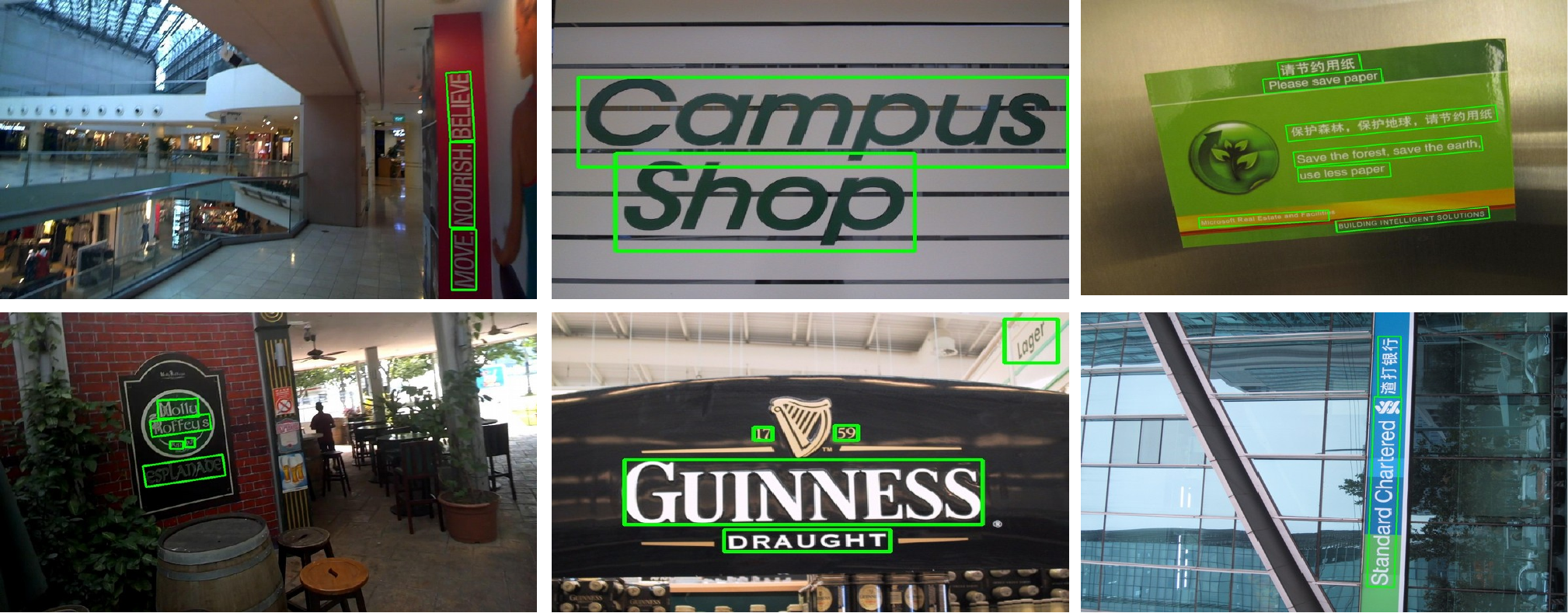}
		\end{center}
		\caption{Examples of detection results. From left to right in columns: IC15, IC13, and MSRA-TD500.}
		\label{fig:result-examples}
	\end{figure*}

	PixelLink models are trained and evaluated on several benchmarks, achieving on par or better results than state-of-the-art methods, showing that the text localization task can be well solved without bounding box regression. Some detection results are shown in \mbox{Fig.~\ref{fig:result-examples}}. 
	\subsection{Benchmark Datasets}
	\subsubsection{ICDAR2015(IC15)}
	Challenge 4 of IC15 ~\cite{Karatzas2015ICDAR} is the most commonly used benchmark for detecting scene text in arbitrary directions. It consists of two sets: train and test, containing $1,000$ and 500 images respectively. Unlike previous ICDAR challenges, images are acquired using Google Glass without taking care of viewpoint, positioning or frame quality. Only readable Latin scripts longer than 3 characters are cared and annotated as word quadrilaterals. `do not care' scripts are also annotated, but ignored in evaluation.
	\subsubsection{ICDAR2013(IC13)}
	IC13~\cite{Karatzas2013ICDAR} is another widely used benchmark for scene text detection, containing 229 images for training, and 233 for testing. Text instances in this dataset are mostly horizontal and annotated as rectangles in words.
	
	\subsubsection{MSRA-TD500(TD500)}
	Texts in TD500 ~\cite{Yao2012TD500} are also arbitrarily oriented, but much longer than those in IC15 because they are annotated in lines. TD500 contains 500 images in total, 300 for training and 200 for testing. Both English and Chinese scripts exist.

	Corresponding standard evaluation protocols are used.
	\subsection{Implementation Details}
	PixelLink models are optimized by SGD with momentum $=0.9$ and weight decay = $5\times10^{-4}$. Instead of fine-tuning from an ImageNet-pretrained model, the VGG net is randomly initialized via the xavier method ~\cite{Glorot2010Xavier}. Learning rate is set to $10^{-3}$ for the first 100 iterations, and fixed at $10^{-2}$ for the rest. Details will be described in each experiment individually.
	
	The whole algorithm is implemented in Tensorflow 1.1.0 and pure Python, with the code of join operation described in Sec.~\ref{sec:linking-pixels-together} compiled with Cython. When trained with a batch size of 24 on 3 GPUs(GTX Titan X), it takes about 0.65s per iteration, and the whole training processing takes about 7$\sim$8 hours. 128G RAM and two Intel Xeon CPUs(2.20GHz) are available on the machine where experiments are conducted.

	\subsection{Detecting Oriented Text in IC15}
	\label{sec:results-on-ic15}
	The training starts from a randomly initialized VGG16 model, on IC15-train only. Model of 4s requires about 40K iterations of training, and 2s longer, about 60K iterations. Minimal shorter side length and area are used for post-filtering and set to 10 and 300 respectively, the corresponding 99th-percentiles of IC15-train by ignoring `do-not-care' instances. Thresholds on pixel and link are found by grid search and set to (0.8, 0.8). Input images are resized to $1280\times 768$ in testing. Results are shown in \mbox{Tab.~\ref{table:results-on-ic15}}.
	\begin{table}
		\centering
		\caption{Results on IC15.
			`R, P, F' stand for Recall, Precision and F-score. All the listed methods are tested at 720P or similar $1280\times 768$. `---' means unreported. `MS' stands for Multi-Scale. PixelLink+VGG16 2s and 4s predict at \emph{conv2\_2} and  \emph{conv3\_3} respectively. PVANET2x is a modified version of PVANET~\cite{kim2016pvanet}, by doubling the channels. All listed results for comparison are quoted from the corresponding original papers.}
		\label{table:results-on-ic15}
		\begin{tabular}{|l|c|c|c|c|}
			\hline
			Model                   &R      &P      &F      &FPS\\
			\hline
			\hline
			\textbf{PixelLink+VGG16 2s}     &\textbf{82.0}  &\textbf{85.5}  &\textbf{83.7}  &3.0\\
			\hline
			\textbf{PixelLink+VGG16 4s}      &81.7   &82.9  &82.3   &7.3\\
			\hline
			EAST+PVANET2x MS       &78.3   &83.3  &81.0   &---\\
			\hline
			EAST+PVANET2x           &73.5   &83.6  &78.2   &\textbf{13.2}\\
			\hline
			EAST+VGG16              &72.8   &80.5  &76.4   &6.5\\
			\hline
			SegLink+VGG16           &76.8   &73.1  &75.0   &---\\
			\hline
			CTPN+VGG16              &51.6   &74.2  &60.9   &7.1\\
			\hline
		\end{tabular}
	\end{table}
	
	The best performance of PixelLink on IC15 is better than the existing best single-scale method (EAST+PVA2x) by 5.5\% in F-score. To further check the generalization ability of PixelLink, the same 2s model is also tested on COCO-Text evaluation set without any fine-tuning, achieving a performance of 35.4, 54.0, 42.4 for recall, precision and F-score, exceeding the corresponding results 32.4, 50.4, 39.5 of EAST~\cite{Zhou2017EAST}.
	
	For the differences on implementation details and running environment, it's not a easy task to conduct an objective and fair speed comparison. Speeds reported here are only intended to show that PixelLink is not slower when compared with the regression-based state-of-the-art methods, if the same deep model, \emph{i.e.}, VGG16, is used as the base network. Note it that although the EAST+PVANET2x runs the fastest, its accuracy is much lower than PixelLink models.

	\subsection{Detecting Long Text in TD500}
	Since lines are detected in TD500, the final model on IC15 is not used for fine-tuning. Instead, the model is pretrained on IC15-train for about 15K iterations and fine-tuned on TD500-train + HUST-TR400 ~\cite{Yao2014TR400} for about 25K iterations. Images are resized to $768 \times 768$ for testing. Thresholds of pixel and link are set to (0.8, 0.7). The min shorter side in post-filtering is 15 and min area 600, as the corresponding 99th-percentiles of the training set.
	
	\begin{table}
		\centering
		\caption{Results on MSRA-TD500.  All listed results for comparison are quoted from the corresponding original papers.}
		\label{table:results-on-msra-td500}
		\begin{tabular}{|l|c|c|c|c|}
			\hline
			Method              &Recall     &Precision      &F-score\\
			\hline
			\hline
			\textbf{PixelLink + VGG16 2s}    &73.2   &83.0   &\textbf{77.8} \\
			\hline
			\textbf{PixelLink + VGG16 4s}    &73.0   &81.1   &76.8 \\
			\hline
			EAST + PVANET2x       &67.4   &\textbf{87.3}  &76.1 \\
			\hline
			EAST + VGG16        &61.6   &81.7   &70.2 \\
			\hline
			SegLink + VGG16     &70.0   &86.0   &77.2 \\
			\hline
			\cite{yao2016scene} &\textbf{75.3}  &76.5   &75.9 \\
			\hline
		\end{tabular}
	\end{table}
	
	Results in Tab.~\ref{table:results-on-msra-td500} show that among all VGG16-based models, EAST behaves the worst for its highest demand on large receptive fields. Both SegLink and PixelLink don't need a deeper network to detect long texts, for their smaller demands on receptive fields.
	\subsection{Detecting Horizontal Text in IC13}
	\label{sec:ic13-result}
	The final models for IC15 are fine-tuned on IC13-train, TD500-train and TR400, for about 10K iterations. In single scale testing, all images are resized to $512\times512$. Multi-scales includes (384, 384), (512, 512), (768, 384), (384, 768), (768, 768), and a maximum longer side of 1600. Thresholds on pixel and link are (0.7, 0.5) for single-scale testing, and (0.6, 0.5) for multi-scale testing. The 99-th percentiles are 10, 300, for shorter side and area respectively, for post-filtering.
	
	Different from regression-based methods like EAST and TextBoxes, PixelLink has no direct output as confidence on each detected bounding box, so its multi-scale testing scheme is specially designed. Specifically, prediction maps of different scales are uniformly resized to the largest height and the largest width among all maps. Then, fuse them by taking the average. The rest steps are identical to single-scale testing.
	
	Results in Tab.\ref{table:results-on-ic13} show that multi-scale leads to an improvement of about 4 to 5 points in F-score, similar to the observation in TextBoxes.
	\begin{table}
		\caption{Results on IC13, in DetEval. `MS' stands for multi-scale testing. `MS?' means it is not reported whether MS is used.  All listed results for comparison are quoted from the corresponding original papers.}
		\label{table:results-on-ic13}
		\begin{tabular}{|l|c|c|c|}
			\hline
			Method                  &Recall &Precision  &F\\
			\hline
			\hline
			\textbf{PixelLink+VGG16 2s}      &83.6   &86.4       &84.5\\ \hline
			\textbf{PixelLink+VGG16 4s}      &82.3   &84.4       &83.3\\ \hline
			\textbf{PixelLink+VGG16 2s} MS   &\textbf{87.5}  &88.6       &\textbf{88.1}\\ \hline
			\textbf{PixelLink+VGG16 4s} MS   &86.5  &88.6       &87.5\\ \hline
			TextBoxes+VGG16         &74     &88         &81  \\ \hline
			TextBoxes+VGG16 MS      &83     &89         &86  \\ \hline
			EAST+PVANET2x MS?          &82.7   &92.6       &87.4\\ \hline
			SegLink+VGG16              &83.0   &87.7       &85.3\\ \hline
			CTPN + VGG16            &83.0 &\textbf{93.0}&87.7\\ \hline
		\end{tabular}
	\end{table}
	
	\section{Analysis and Discussion}
	\subsection{The Advantages of PixelLink}
	\label{sec:seg-vs-reg}
	\begin{table*}[!ht]
		\caption{Comparison of training speed and training data for IC15. SynthText~\cite{Gupta2016SynthText} is a synthetic dataset containing more than 0.8M images. All listed results for comparison are quoted from the corresponding original papers.}
		\label{table:cmp-between-seg-and-reg}
		\centering
		\begin{tabular}{|l|c|c|c|c|c|c|c|}
			\hline
			Method                          &ImageNet Pretrain  &Optimizer  &Training Data              &Iterations & F-score       \\
			\hline\hline
			\textbf{PixelLink+VGG16 4s} at Iter-25K  &No                     &SGD        &IC15-train                 &$\simeq$25K        &79.7   \\
			\hline
			\textbf{PixelLink+VGG16 4s} at Iter-40K  &No                     &SGD        &IC15-train                 &$\simeq$40K        &82.3   \\
			\hline
			SegLink + VGG16                 &Yes                    &SGD        &SynthText,IC15-train       &$\simeq$100K       &75.0   \\
			\hline
			EAST+VGG16                      &Yes                    &ADAM       &IC15-train,IC13-train      &$>$55K             &76.4   \\
			\hline
		\end{tabular}
	\end{table*}
	A further analysis of experiment results on IC15 shows that PixelLink, as a segmentation based method, has several advantages over regression based methods.  As listed in  Tab.~\ref{table:cmp-between-seg-and-reg}, among all methods using VGGNet, PixelLink can be trained much faster with less data, and behaves much better than the others. Specifically, after about only 25K iterations of training (less than a half of those needed by EAST or SegLink), PixelLink can achieve a performance on par with SegLink or EAST. Keep in mind that, PixelLink is trained from scratch, while the others need to be fine-tuned from an ImageNet-pretrained model. When also trained from scratch on IC15-train, SegLink can only achieve a F-score of 67.8\footnote{This experiment is repeated  3 times using the open source code of SegLink. Its best performance is 63.6, 72.7, 67.8 for Recall, Precision and F-score respectively.}.

	The question arises that, why PixelLink can achieve a better performance with many fewer training iterations and less training data?  We humans are good at learning how to solve problems. The easier a problem is, the faster we can learn, the less teaching materials we will need, and the better we are able to behave. It may also hold for deep-learning models. So two factors may contribute.
	
	\subsubsection{Requirement on receptive fields}When both adopting VGG16 as the backbone, SegLink behaves much better than EAST in long text detection, as shown in Tab.~\ref{table:results-on-msra-td500}. This gap should be caused by their different requirements on receptive fields. Prediction neurons of EAST are trained to observe the whole image, in order to predict texts of any length. SegLink, although regression based too, its deep model only tries to predict segments of texts, resulting in a less requirement on receptive fields than EAST.
	
	\subsubsection{Difficulty of tasks}In regression-based methods, bounding boxes are formulated as quadrangle or rotated rectangle. Every prediction neuron has to learn to predict their locations as precise numerical values, \emph{i.e.} coordinates of four vertices, or center point, width, height and rotation angle. It's possible and effective, as has long been proven by algorithms like Faster R-CNN, SSD, YOLO, \emph{etc.} However, such kind of predictions is far from intuitive and simple. Neurons have to study a lot and study hard to be competent for their assignments.
	
	When it comes to PixelLink, neurons on the prediction layer only have to observe the status of itself and its neighboring pixels on the feature map. In another word, PixelLink has the least requirement on receptive fields and the easiest task to learn for neurons among listed methods.
	
	A useful guide for the design of deep models can be picked up from the assumption above: simplify the task for deep models if possible, because it might make them get trained faster and less data-hungry.
	
	The success in training PixelLink from scratch with a very limited amount of data also indicates that text detection is much simpler than general object detection. Text detection may rely more on low-level texture feature and less on high-level semantic feature.
	
	\subsection{Model Analysis}
	\label{sec:model-analysis}
	As shown in Tab.~\ref{table:ablation-results}, ablation experiments have been done to analyze PixelLink models.
	Although PixelLink+VGG16 2s models have better performances, they are much slower and less practical than the 4s models, and therefore, Exp.1 of the best 4s model is used as the basic setting for comparison.
	\begin{table}[!ht]
		\caption{Investigation of PixelLink with different settings on IC15. `R': Recall, `P': Precision; `F': F-score.
		}
		\label{table:ablation-results}
		\centering
		\begin{tabular}{|c|c|c|c|c|c|c|c|c|c|}
			\hline
			\#      &Configurations &R      &P      &F    \\
			\hline
			\hline
			1       &The best 4s model  &81.7   &82.9   &82.3 \\
			\hline
			2       &Without link mechanism &58.0   &71.4   &64.0 \\
			\hline
			3       &Without Instance Balance  &80.2   &82.3   &81.2 \\
			\hline
			4       &Training on 384x384 &79.6   &81.2   &80.4 \\
			\hline
			5       &No Post-Filtering &82.3   &52.7   &64.3 \\
			\hline
			6       &Predicting at 2s resolution &82.0   &85.5   &83.7 \\
			\hline
		\end{tabular}
	\end{table}
	\subsubsection{Link is very important.}
	In Exp.2, link is disabled by setting the threshold on link to zero, resulting in a huge drop on both recall and precision. The link design is important because it converts semantic segmentation into instance segmentation, indispensable for the separation of nearby texts in PixelLink.
	\subsubsection{Instance-Balance contributes to a better model.}
	In Exp.3, Instance-Balance (IB) is not used and the weights of all positive pixels are set to the same during loss calculation. Even without IB, PixelLink can achieve a F-score of 81.2, outperforming state-of-the-art. When IB is used, a slightly better model can be obtained. Continuing the experiments on IC13, the performance gap is more obvious(shown in Tab.~\ref{table:ib-on-ic13}).  
	\begin{table}
		\caption{Effect of Instance-Balance on IC13. The only difference on the two settings is the use of Instance-Balance. Testing is done without multi-scale.}
		\label{table:ib-on-ic13}
		\centering
		\begin{tabular}{|c|c|c|c|}
			\hline
			IB  &Recall     &Precision      &F-score \\ \hline \hline
			Yes &82.3       &84.4           &83.3 \\ \hline
			No  &79.4       &84.2           &81.7 \\ \hline
		\end{tabular}
	\end{table}
	\subsubsection{Training image size matters.}
	In Exp.4, images are resized to $384\times 384$ for training, resulting in an obvious decline on both recall and precision. This phenomenon is in accordance with SSD.
	\subsubsection{Post-filtering is essential.}
	In Exp.5, post-filtering is removed, leading to a slight improvement on recall, but a significant drop on precision.
	\subsubsection{Predicting at higher resolution is more accurate but slower.}
	In Exp.6, the predictions are conducted on \emph{conv2\_2}, and the performance is improved \emph{\mbox{w.r.t.}} both recall and precision, however, at a cost of speed. As shown in Tab.~\ref{table:results-on-ic15}, the speed of 2s is less than a half of 4s, demonstrating a tradeoff between performance and speed.
	
	\section{Conclusion and Future Work}
	PixelLink, a novel text detection algorithm is proposed in this paper. The detection task is achieved through instance segmentation by linking pixels within the same text instance together. Bounding boxes of detected text are directly extracted from the segmentation result, without performing location regression. Since smaller receptive fields are required and easier tasks are to be learned, PixelLink can be trained from scratch with less data in fewer iterations, while achieving on par or better performance on several benchmarks than state-of-the-art methods based on location regression.
	
	VGG16 is chosen as the backbone for convenient comparisons in the paper. Some other deep models will be investigated for better performance and higher speed.
	
	Different from current prevalent instance segmentation methods~\cite{Li2016FCIS}~\cite{He2017MaskRCNN}, PixelLink does not rely its segmentation result on detection performance. Applications of PixelLink will be explored on some other tasks that require instance segmentation.
	\section*{Acknowlegement}
	This work was supported by National  Natural  Science  Foundation  of  China  under
	Grant 61379071. Special thanks to Dr. Yanwu Xu in CVTE Research for all his kindness and great help to us.
	\bibliographystyle{aaai}
	\bibliography{ref}
	
\end{document}